\documentclass[letterpaper, 10 pt, conference]{ieeeconf}
\usepackage{graphics} 
\usepackage{epsfig} 
\usepackage{mathptmx} 
\usepackage{times} 
\usepackage{amsmath} 
\usepackage{amssymb}  
\usepackage{bm} 
\usepackage{cite}
\usepackage{url}
\usepackage{color}

\title{\LARGE \bf Design of a Variable Stiffness Spring with Human-Selectable Stiffness}

\author{Chase W. Mathews and David J. Braun
\thanks{C. W. Mathews and D. J. Braun are with the Advanced Robotics and Control Laboratory within the Center for Rehabilitation Engineering and Assistive Technology, Department of Mechanical Engineering, Vanderbilt University, Nashville, Tennessee 37235, USA.
\newline
\indent This work was supported in part by a Seeding Success Grant provide by Vanderbilt University and a National Science Foundation CAREER Award (Grant No. 2144551). The authors gratefully acknowledge the support.}
\thanks{E-mail: {\tt\small chase.w.mathews@vanderbilt.edu}}
\thanks{ E-mail: {\tt\small david.braun@vanderbilt.edu}}
}
\IEEEoverridecommandlockouts
\begin{document}
\maketitle
\begin{abstract}
Springs are commonly used in wearable robotic devices to provide assistive joint torque without the need for motors and batteries. However, different tasks (such as walking or running) and different users (such as athletes with strong legs or the elderly with weak legs) necessitate different assistive joint torques, and therefore, springs with different stiffness. Variable stiffness springs are a special class of springs which can exert more or less torque upon the same deflection, provided that the user is able to change the stiffness of the spring. In this paper, we present a novel variable stiffness spring design in which the user can select a preferred spring stiffness similar to switching gears on a bicycle. Using a leg-swing experiment, we demonstrate that the user can increment and decrement spring stiffness in a large range to effectively assist the hip joint during leg oscillations. Variable stiffness springs with human-selectable stiffness could be key components of wearable devices which augment locomotion tasks, such as walking, running, and swimming.
\end{abstract}
\section{Introduction}
Mechanical springs are commonly used in wearable devices \cite{Sawicki2020} to provide assistive torque without the use of motors and batteries.
Prior works have shown that unpowered spring-driven exoskeletons can reduce the metabolic cost of walking by $7\%$ \cite{Collins2015} and running by $8\%$ \cite{Nasiri2018}; in both cases, a fixed stiffness spring was optimized across users.
Alternatively, a variable stiffness spring with selectable stiffness could enable users to choose the most optimal spring stiffness for the task, similar to how a bicycle derailleur enables cyclists to select the most optimal gear ratio independent of the cyclist and the cycling speed.

Variable stiffness springs have been previously used in lower limb prostheses \cite{Tran2019,Shepherd2017} and orthoses \cite{Weinberg2007,Thalman2019} to help humans with motions such as walking, running, and stair-ascent. 
Many of the previously designed variable stiffness springs \cite{Jafari2014,Lau2018,Braun2019,Braun2019b,Mathews2021} ensure that a small force can be used to adjust the spring stiffness when the spring is unloaded \cite{Chalvet2017a}.
However, for oscillatory motions such as the swing of the hip during walking, running, or swimming, the spring may only be at equilibrium for a fraction of a second during each cycle of the motion.
Consequently, if the human aims to effortlessly change stiffness during an oscillatory task, then the human must precisely apply the force to change stiffness when the spring is not deflected at each cycle.

Using a variable stiffness robot actuator, we have previously shown that a small but fast motor can change spring stiffness during oscillatory motion, once each motion cycle, by applying precisely timed forces \cite{Mathews2021}. Replacing the motor with a human limb and a bicycle hand shifter is impractical because the user would be required to operate the hand shifter at precise times. 
However, by placing a spring in series between the hand shifter and the mechanism which adjusts spring stiffness, the requirement for precisely timed movements can be eliminated. We use this idea of replacing a weak but fast motor with a human finger in our design of a variable stiffness spring with human-selectable stiffness.
\begin{figure}[t!]
	\centering
	\includegraphics[width=\columnwidth]{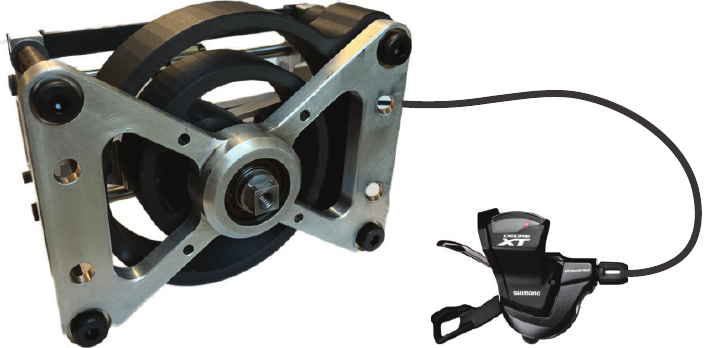}
	\caption{Variable stiffness spring with human-selectable stiffness.}
	\label{IntroFigure}
\end{figure} 

In this paper, we propose the design of a variable stiffness mechanism which enables the human to change the stiffness of the spring in the same way the derailleur enables cyclists to change the gear ratio of the bicycle.
The device consists of a 3D printed composite spiral spring and a variable stiffness mechanism. The stiffness of the spring is changed by a series-spring actuated Bowden cable hand shifter, and a unique self-locking mechanism implemented with a linear ratchet and pawl. 
We show that the device allows the user to effectively change the assistance provided to the user in a leg swing experiment, where the hip joint of the human is augmented with the variable stiffness spring. We further demonstrate that the user can increment and decrement the spring stiffness using a bicycle hand shifter, the same way a cyclist would down-shift or up-shift the gear ratio to adapt to different terrains and speeds while riding the bicycle. 

\section{Model} \label{model}
\begin{figure}[t]
	\centering
	\includegraphics[width=\columnwidth]{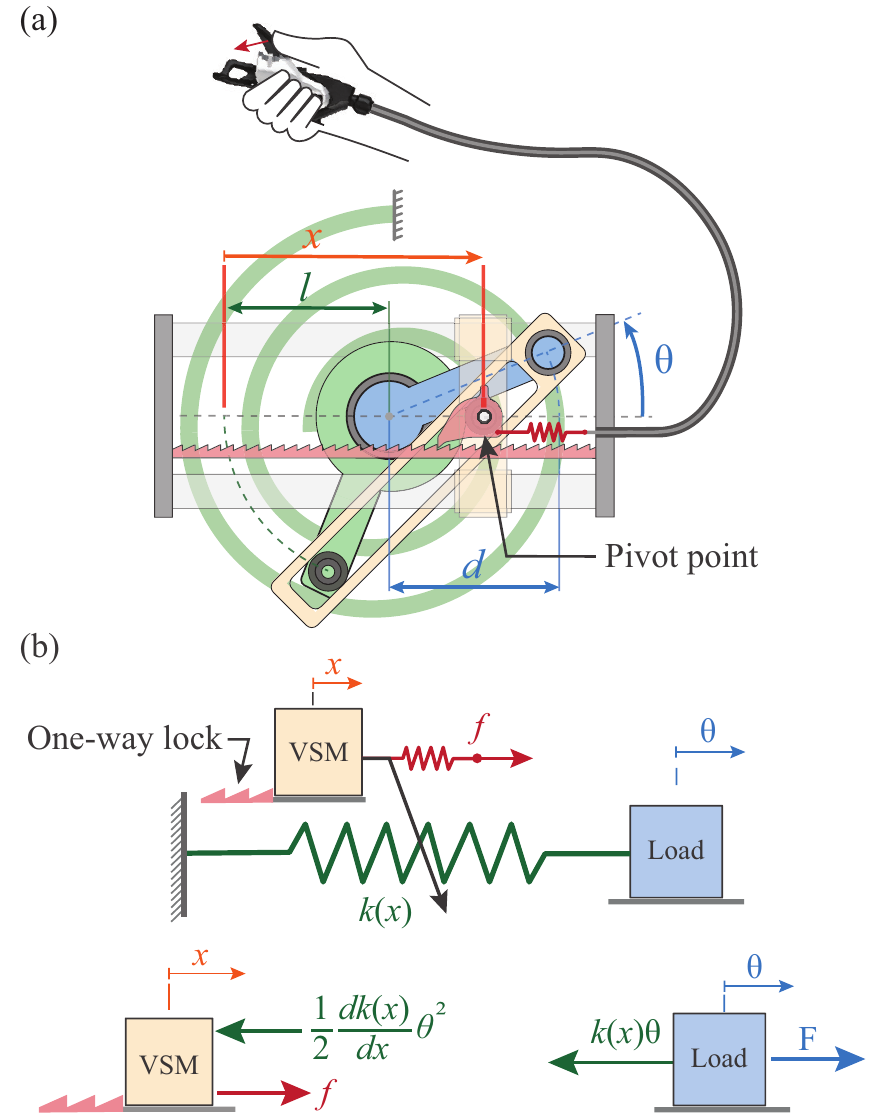}
	\caption{(a) Model of the variable stiffness spring with human-selectable stiffness, consisting of the hand shifter, series spring, and the self-locking variable stiffness mechanism. The hand shifter enables the human to extend the series spring, which changes the position of the pivot point $x$. (b) Schematic representation of the variable stiffness spring.}
	\label{ShifterMechanism}
\end{figure}

The model of the human-driven variable stiffness spring joint is shown in Fig.~\ref{ShifterMechanism}a. The joint is composed of a spring and a mechanism that changes the stiffness of the spring. The free body diagram of the model is shown in Fig.~\ref{ShifterMechanism}b. Below we present the model of the mechanism.

The torque-angle relation of the variable stiffness spring is defined by (see Fig.~\ref{ShifterMechanism}b):
\begin{align}
\label{tau}
\tau = \tau(x,\theta) \approx k(x) \theta,
\end{align}
where $\tau$ is the joint torque, $\theta$ is the joint angle, $k$ is the joint stiffness, while $x$ denotes the position of the pivot point.

The joint stiffness $k(x)$ depends on the design of the mechanism. The stiffness of the torsional spring introduced in \cite{Mathews2021}, and shown in Fig.~\ref{ShifterMechanism}a, is given by the following relation:
\begin{align}
	k(x)= \frac{\partial \tau}{\partial \theta}\bigg|_{\theta=0}
	\approx k_S\bigg(\frac{x}{l+d-x}\bigg)^2,
	\label{StiffnessEquation}
\end{align} 
where $k_S$ is the constant stiffness of the torsional spring attached to the joint, $d$ is the length of the linkage attached to the shaft, while $l$ is the length of the linkage attached to the spring. According to (\ref{StiffnessEquation}), the stiffness of the joint is a monotonically increasing function of $x$.

Changing the position of the pivot point changes the mechanical advantage of the spring over the joint. The equation governing the motion of the pivot point is given by: 
\begin{align}
\label{x}
m\ddot{x} = f - F(\theta,x) = f -\frac{1}{2}\frac{dk(x)}{dx}\theta^2,
\end{align}
where $m$ is the mass of the stiffness modulating mechanism, $f$ is the external force applied to modulate joint stiffness, while $F$ is the reaction force of the spring aiming to move the pivot point by back-driving the stiffness modulating mechanism (see Fig.~\ref{ShifterMechanism}b). The latter effect appears because the spring always tends to move the pivot-point to the position associated with the lowest joint stiffness.

In order to prevent the reaction force of the spring from changing the position of the pivot-point, we use a spring-loaded linear ratchet-pawl mechanism (Fig.~\ref{ShifterMechanism}a). The mechanism locks when the force required to move the pivot point $f$ exceeds the threshold locking force $f_0$ of the spring loaded pawl; it is unlocked otherwise. 

When the ratchet-pawl mechanism is locked, the reaction force of the joint spring cannot move the pivot point in the direction that lowers the joint stiffness,
\begin{align}
\label{xdot}
&0<f_0 < f \Rightarrow  \dot{x} \geq 0,
\end{align}
but the externally applied force $f$ can be used to move the pivot point and increase the joint stiffness, given that the externally applied force is larger than the reaction force of the spring, $f > F(\theta,x)$ in (\ref{x}).

When the ratchet-pawl mechanism is unlocked $0< f < f_0$, the reaction force of the spring can be used to lower the joint stiffness under the following condition $0< f < f_0 < F(\theta,x)$ in (\ref{x}). This condition will be satisfied when the spring is considerably deflected, as in that case, the reaction force of the spring is much larger than the threshold force $f_0$ used to lock the ratchet-pawl mechanism.

\section{Working Principle} \label{concept}
We will now examine the physical requirements to change the joint stiffness using small forces.

Let us consider a variable stiffness joint placed at the human hip and attached to the leg. Let us further assume that the leg swings back and forth with frequency $\omega$, and amplitude $\theta_{\max}$, in walking or running, 
\begin{align}
	\theta = \theta_{\max}\sin \omega t.
	\label{Motion}
\end{align}

\begin{figure*}[ht]
	\centering
	\includegraphics[width=\textwidth]{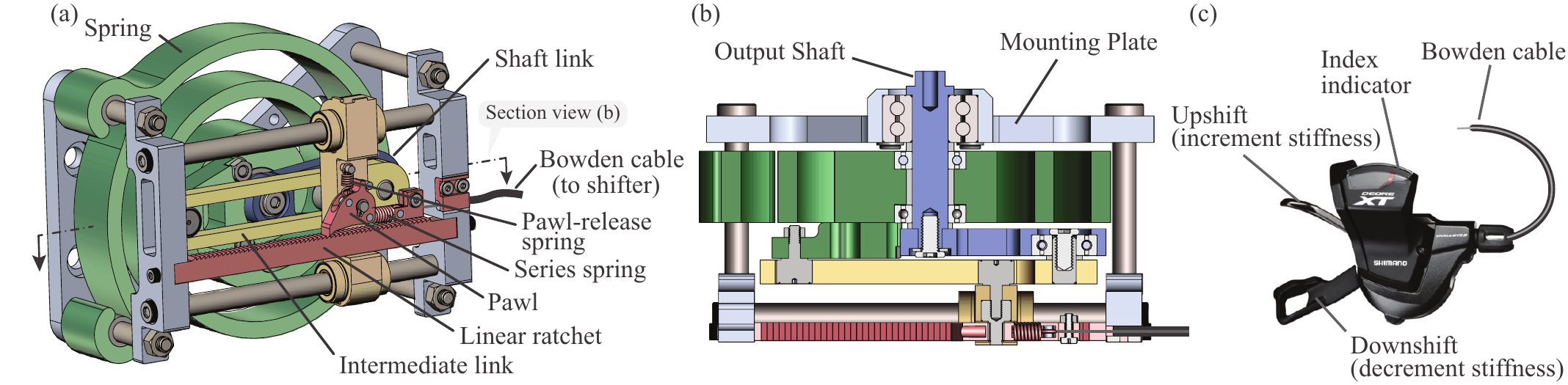}
	\caption{Design of the spring mechanism. (a) Isometric view. (b) Section view. (c) Bowden cable and the hand shifter. The size of the mechanism is 150mm x 130mm x 77mm. The mass of the mechanism is 1.13kg. The combined mass of the hand shifter and the Bowden cable is 0.3kg.}
	\label{Mechanism}
\end{figure*}

Now, to increase the joint stiffness, we must apply an external force $f$ that is larger than the spring force opposing the motion of the pivot point,
\begin{equation}
f>  F(\theta,x) = \frac{1}{2}\frac{dk(x)}{dx}\theta^2  \Rightarrow  \ddot{x} > 0.
\label{Lock}
\end{equation}
This condition can be satisfied by an arbitrarily small force $f$, when the spring is un-deflected, $\theta = 0$, or a small force when the spring is slightly deflected, $\theta \approx 0$. 

Now, let us assume that the force provided by the human is limited,
\begin{equation}
	0\leq f\leq f_{\max},
	\label{LockLimits}
\end{equation}
and using (\ref{Motion}) and (\ref{Lock}), let us estimate the time available to change the stiffness of the joint using the maximal force that can be provided by the human
\begin{align}
\frac{\Delta t}{T} < \frac{2}{\pi}\sqrt{\frac{f_{\max}}{F_{\max}}},
	\label{timing}
\end{align}
where $T=2\pi/\omega$ is the period of leg oscillations. 

Relation (\ref{timing}) suggests that if the maximal force to change the spring stiffness $f_{\max}$ is small compared to the maximal reaction force of the spring $F_{\max}$, then the time window in which the joint stiffness can be changed $\Delta t$ is also small:
\begin{align}
\frac{f_{\max}}{F_{\max}} \ll 1 \;\; \Rightarrow \;\; \frac{\Delta t}{T} \ll 1.
	\label{timing0}
\end{align}
Therefore, the timing of the external force $f$ must be precise in order to increase the spring stiffness with a small force.

The requirement for precise timing is circumvented in our device by using a spring between the hand shifter and the pivot point, see Fig.~\ref{ShifterMechanism} (series spring). 

In a typical work cycle, the human would actuate the hand shifter once or multiple times in order to reduce the length of the Bowden cable, and consequently extend the series spring of stiffness $k_s$, until the maximum force is reached,
\begin{equation}
f = k_s \Delta x \leq f_{\max}.
\label{smallspringstiffness}
\end{equation}

If the ratchet-pawl mechanism is unlocked, then extending the spring will move the pivot point and increase the joint stiffness. If the ratchet-pawl mechanism is locked, extending the spring will increase the force in the series spring but will not move the pivot point or increase the joint stiffness. 
Since the series spring can be extended while the ratchet-pawl mechanism is locked (\ref{Lock}), the human can use the hand shifter over nearly the whole period of oscillations $T$, except perhaps the short time window $\Delta t$ when the ratchet-pawl mechanism unlocks and the series spring moves the pivot point to increase the joint stiffness 
\begin{align}
	\frac{T-\Delta t}{T} \approx 1.
	\label{timing1}
\end{align}
Consequently, the time available for the human to apply force is not limited to $\Delta t$, and is largely independent of the force applied by the human.

In summary, the series spring removes the precise timing requirement to change the joint stiffness using small forces. The series spring also enables the user to extend the spring over multiple oscillation cycles, which further mitigates the precise timing required to change the joint stiffness without the series spring (\ref{timing0}), or the time available to change stiffness during a single cycle with the series spring (\ref{timing1}).
\section{Prototype} \label{design}
In this section, we show the design of the human-selectable variable stiffness spring. Figure~\ref{Mechanism} shows the main components of the mechanism; (i) the torsional variable stiffness spring, (ii) the hand shifter, and (iii) the self-locking pivot point mechanism.
In the following we describe each of these components in detail.
\subsection{The torsional variable stiffness spring}
We use a 3D printed carbon fiber reinforced spiral torsion spring (Fig.~\ref{Mechanism} green spring) due to customizability and large volumetric energy density. 
The spring was fabricated using Onyx (nylon filled with chopped carbon fiber)
with half of the layers reinfored with continuous strands of carbon fiber.
The carbon fiber spring has an estimated stiffness of $k_S \approx 24 ~\text{Nm/rad}$.
We use the adjutable pivot-point linkage mechanism presented in \cite{Mathews2021} (Fig.~\ref{Mechanism} yellow) to vary the mechanical advantage of the spiral spring over the joint. 

\begin{figure}[ht!]
	\centering
	\includegraphics[width=1\columnwidth]{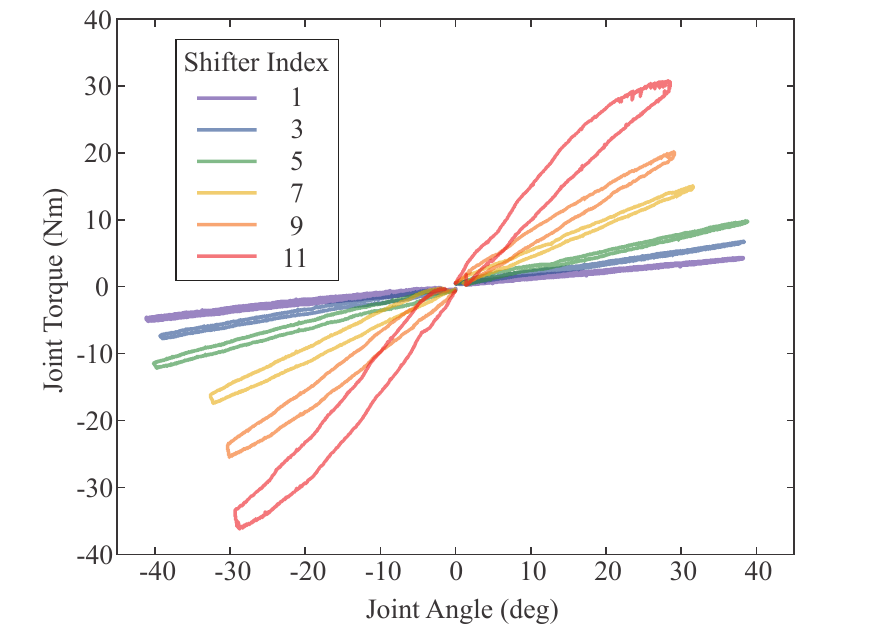}
	\caption{Torque-angle characteristics of the variable stiffness spring.}
	\label{TorqueDeflection}
\end{figure}

We performed a static deflection experiment to estimate the torque-deflection behavior of the variable stiffness joint. During the experiment, the spring was deflected at six different stiffness configurations. The measured torque-angle data is shown in Fig.~\ref{TorqueDeflection}. 

\begin{figure*}[t!]
	\centering
	\includegraphics[width=\textwidth]{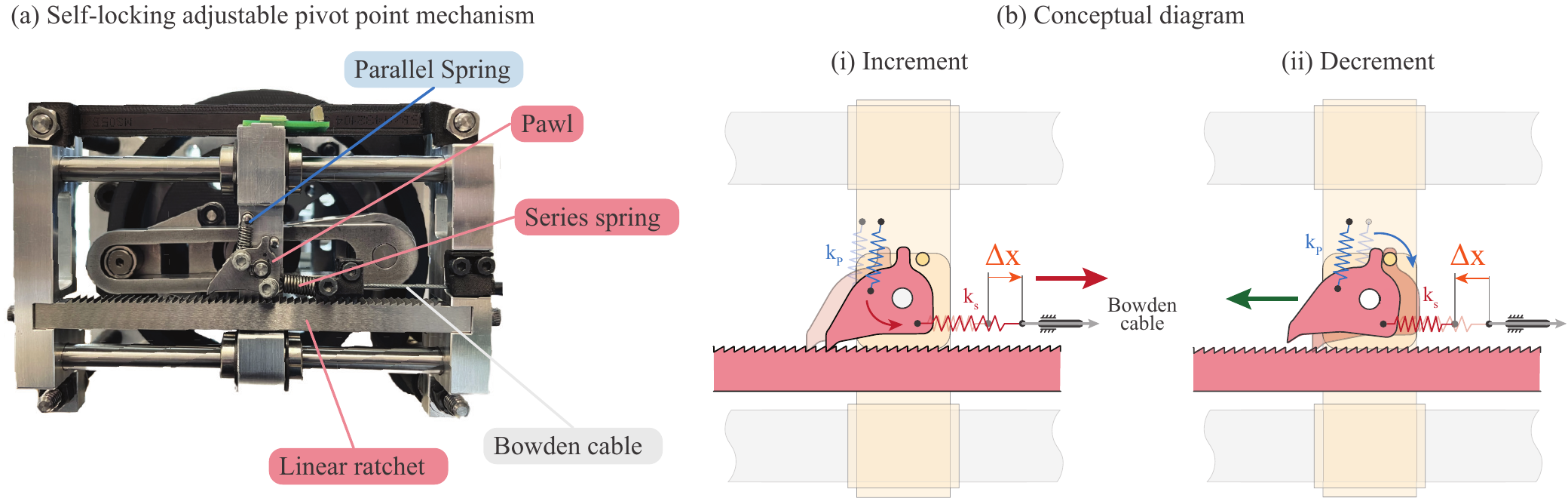}
	\caption{(a) Prototype showing the self-locking pivot point mechanism. (b) Working principle of the self-locking pivot point mechanism; (i) the configuration for increasing stiffness, and (ii) the configuration for decreasing stiffness. The supplementary video shows the operation of the mechanism.}
	\label{VSSDiagram}
\end{figure*}

The joint is characterized with the following torsional stiffness values,
\begin{align}
	\label{ars}
	k\in[6,70] ~\text{Nm/rad}.
\end{align}
This stiffness enables the spring to provide $\pm[3,36]$~Nm assistive joint torque at $\pm 30$~deg joint rotation, which amounts to around $35\%$ of the average human hip torque for a $75$~kg person during walking at normal speeds ($1.6$~m/s) \cite{Grimmer2014}.

Finally, Fig.~\ref{TorqueDeflection} also shows that the hand shifter and the locking mechanism can be used to change the stiffness and hold the torque provided by the torsional spring.

\subsection{The hand shifter}
The hand shifter is shown in Fig.~\ref{Mechanism}c.
The hand shifter has two levers that, when pressed, either extend or retract the Bowden cable.
The Bowden cable is a steel cable routed inside a flexible housing.
In the bicycle, the Bowden cable shifts the chain between different sets of sprockets.

Bowden cables have been extensively used in wearable devices to transfer force \cite{Wu2015,Cappello2016,Chiu2021}.
In our device, the two ends of the Bowden cable are fastened to the hand shifter and the series spring, while the series spring is connected to the pawl which can rotate about the pivot point, as shown in Fig.~\ref{Mechanism}a. By changing the length of the Bowden cable, the human generates a force and a moment about the pivot point. The moment generated about the pivot point disengages the pawl from the ratchet, such that the force can move the pivot point. In this way, using the hand shifter and the Bowden cable, the human can change the stiffness of the joint. 
\subsection{Self-locking pivot point} \label{PivotPointDesign}
In our device, the pivot point self-locks through the use of a linear ratchet-pawl mechanism, shown in Fig.~\ref{VSSDiagram}.
The mechanism which changes stiffness consists of a linear ratchet rack, a pawl, a series spring between the pawl and the Bowden cable, and a parallel spring between the pawl and the pivot-point housing.
The pawl engages with the ratchet to prevent the motion of the pivot point when the large spiral spring is loaded.
The series and parallel springs of stiffness $k_s$ and $k_p$ (see Fig.~\ref{VSSDiagram}) control the engagement of the pawl to enable the user to change the stiffness of the joint. 

To increase stiffness, the user generates a force on the Bowden cable, which deflects the series spring. 
When the large spiral spring is un-deflected, the series spring pulls the pawl and consequently translates the pivot point (see Fig.~\ref{VSSDiagram}b-i).
The stiffness of the series spring $k_s\approx 6000$~N/m is chosen small enough such that the user can extend it, but large enough such that it can move the pivot point when the large spiral spring is unloaded.

To decrease stiffness, the user extends the Bowden cable, which causes the cable to slack.
When the cable slacks, the series spring becomes unloaded, which allows the parallel spring to disengage the pawl.
In this case, the force imposed by the large deflected spiral spring can move the pivot point, to reduce the joint stiffness, until the slack of the Bowden cable is removed and the pawl is re-engaged (Fig.~\ref{VSSDiagram}b-ii).
The stiffness of the parallel spring $k_p \approx 485$~N/m has been chosen large enough to disengage the pawl around zero joint deflection and small enough to allow the series spring to re-engage the pawl upon a small joint deflection.
\section{Evaluation} \label{experiments}
\begin{figure*}[ht!]
	\centering
	\includegraphics[width=\textwidth]{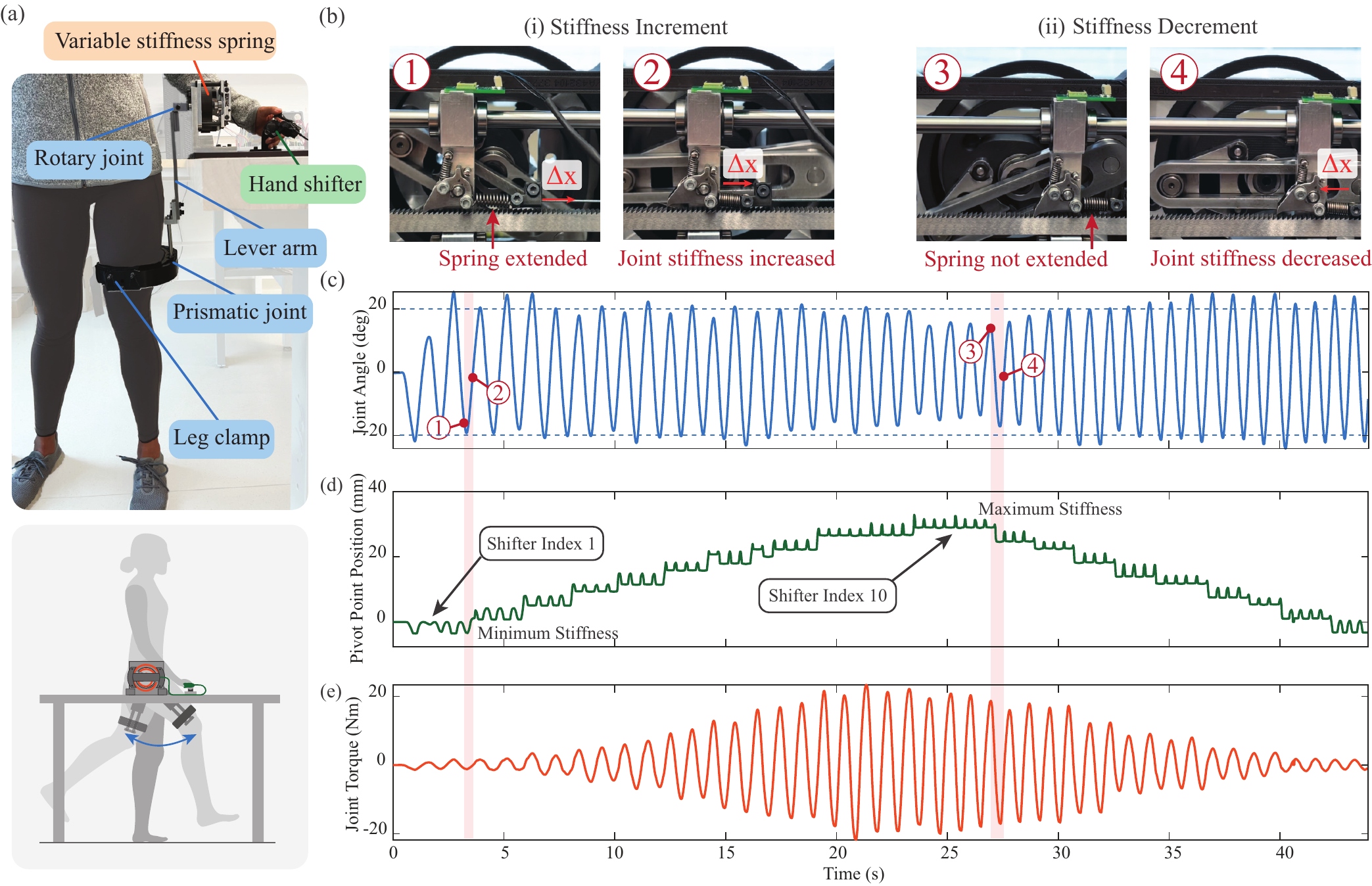}
	\caption{Experiment. (a) Setup. (b) Snapshots of the mechanism while (i) stiffness is incremented and (ii) stiffness is decremented. (c) Angular position of the leg. (d) Position of the pivot point. (e) Estimated torque on the joint.}
	\label{Results}
\end{figure*}

In this section, we present a leg swing experiment where the human-selectable variable stiffness spring joint (shown in Fig.~\ref{IntroFigure}) is used to augment the human hip joint (shown in Fig.~\ref{Results}). The purpose of the experiment is to validate the working principle of the device which enables the human to quickly change joint stiffness over a large range during continuous oscillations.

In what follows, we (i) present the experimental setup, (ii) describe the experimental protocol, and (iii) summarize the results of the experiment.
\subsection{Setup}
To verify the function of the device, we designed a desk-mounted setup where the variable stiffness spring joint can be attached to the leg of a human, as shown in Fig.~\ref{Results}a. 

The leg of the subject was fastened to the spring using a 3D printed clamp, while the clamp was connected to the joint using a lever arm. The lever arm features a rotary joint and a prismatic joint such that the leg is not constrained to move in the lateral and longitudinal directions while being attached to the joint.

The hip angle $\theta$ was captured using a rotary magnetic encoder.
The position of the pivot point $x$ was measured using a linear magnetic encoder. 
The torque provided by the variable stiffness joint was estimated using the the measured joint angle $\theta$, the measured pivot point position $x$, and the experimental torque-deflection curves shown in Fig.~\ref{TorqueDeflection}.

\subsection{Protocol}
We performed a simple exploratory experiment where the subject was asked to (i) swing one leg continuously back and forth at a comfortable frequency and a relatively constant amplitude of $\pm 20$ degrees, (ii) fully increment the bicycle shifter until the maximum stiffness has been reached, and subsequently, (iii) fully decrement the shifter until the minimum stiffness was reached. The protocol was approved by the Institutional Review Board of Vanderbilt University Medical Center (N220192). 

\subsection{Results}
The experimental data is shown in Fig.~\ref{Results}. The mechanism is shown in Fig.~\ref{Results}b while the joint stiffness was incremented (i) and decremented (ii). The angle of the hip joint is shown in Fig.~\ref{Results}c. The position of the pivot point is shown in Fig.~\ref{Results}d. The joint torque is shown in Fig.~\ref{Results}e.

Figure~\ref{Results}b shows snapshots of the device during the experiment with the corresponding timestamps labeled in Fig.~\ref{Results}c. The left snapshot shows an example of a stiffness increment where the series spring is first extended while the leg is away from equilibrium; in this case, the mechanism is locked, and subsequently, the series spring pulls the pivot point to a higher stiffness configuration when the leg is around equilibrium (Fig.~\ref{Results}b-i). The right snapshot shows an example of a stiffness decrement where the Bowden cable is slacked and the parallel spring lifts the pawl until the pivot point moves to a lower stiffness configuation and the slack of the Bowden cable is removed (Fig.~\ref{Results}b-ii).

Figure~\ref{Results}c shows the joint angle during the experiment. We observe that the subject was able to generate continuous oscillatory leg motion such that the amplitude of the joint angle was around $\pm 20$ degrees.

Figure~\ref{Results}d shows the position of the pivot point, and therefore the joint stiffness, during the experiment. We observe that the stiffness was increased from the minimum value (shifter index 1, where $k(x)\approx 6$~Nm/rad) to the maximum value (shifter index 10, where $k(x)\approx 70$~Nm/rad) during the oscillatory motion. 
We note that position of the pivot point oscillates approximately $\pm 2$~mm for each stiffness configuation. This oscillation was caused by the return spring which by default disangages the pawl in order to decrement stiffness as described in Section \ref{PivotPointDesign}. The oscillations did not cause detrimental effects during the experiment.

Figure~\ref{Results}e shows the estimated joint torque of the device during the experiment. We observe that, as the user increased the joint stiffness of the spring, the torque of the spring was also increased from $1$~Nm to around $20$~Nm. 

In summary, the experiment demonstrated that, (i) the locking mechanism was able to successfully hold a given joint stiffness configuration as the device generated $1-20$~Nm joint torque, and (ii) the user was able to effectively use the hand shifter to increment and decrement joint stiffness in a relatively large range of $6-70$~Nm/rad, during continuous swings. The results show that the proposed device enables the human to adapt joint stiffness using an intuitive interface provided by the hand shifter of the bicycle.
\section{Conclusion} \label{discussion}
In this paper, we presented a novel human-adjustable variable stiffness joint in which the user can select different joint stiffness in the same way cyclists select different gear ratios on a bicycle. We show that the device enables the human to use small and non-precisely timed forces to change joint stiffness, rather than requiring the user to provide large forces, or small but precisely timed forces. We verify the ability of the human-adjustable variable stiffness joint to maintain and change stiffness during continuous oscillatory motion. 

Customization of robot exoskeleton assistance for different users, tasks, and speeds has been shown to reduce metabolic demand in walking and running \cite{Sawicki2020,Shepherd2022,Ingraham2022}, and may be useful for other everyday tasks. 
Individualized joint stiffness values could provide benefits to users across different speeds of walking \cite{Shafer2022}.
Individualized joint stiffness values may also help improve joint motion patterns and correct a reduced joint motion range of users with impairment \cite{Zhang2022}.
Finally, individualized joint stiffness values may be used to better augment users performing physically demanding tasks, such as lifting, jumping \cite{Sutrisno2019}, running \cite{Sutrisno2020}, or walking with a heavy load \cite{Zhang2022a}. 

Human-adjustable variable stiffness joints can be key components of mechanically adaptive robot exoskeletons where different users can choose between different levels of assistance for different locomotion tasks.
\bibliographystyle{ieeetr}
\bibliography{bibliographyentries}{}
\end{document}